\pdfoutput=1

\documentclass[11pt]{article}

\usepackage{ACL2023}
\usepackage[inline]{enumitem}
\usepackage{times}
\usepackage{latexsym}
\usepackage[T1]{fontenc}
\usepackage[utf8]{inputenc}
\usepackage{microtype}
\usepackage{inconsolata}
\usepackage{amsmath}
\usepackage{amsfonts}
\usepackage{enumitem}
\usepackage{graphicx}
\usepackage{multirow,adjustbox,makecell,arydshln} 

\renewcommand{\vec}[1]{\boldsymbol{\mathbf{#1}}}

\DeclareMathOperator{\MLP}{MLP}
\DeclareMathOperator{\Softmax}{\sigma}
\DeclareMathOperator{\KL}{KL}
\DeclareMathOperator{\RL}{RL}

\DeclareMathOperator{\gin}{GINConv}
\DeclareMathOperator{\bn}{BN}
\DeclareMathOperator{\relu}{ReLU}
\DeclareMathOperator{\linear}{Linear}
\DeclareMathOperator{\drop}{Dropout}
\DeclareMathOperator{\softplus}{Softplus}
\DeclareMathOperator{\IRBO}{IRBO}
\DeclareMathOperator{\RBO}{RBO}

\title{GINopic: Topic Modeling with Graph Isomorphism Network}

\author{Suman Adhya \and Debarshi Kumar Sanyal \\
    Indian Association for the Cultivation of Science, Jadavpur, Kolkata-700032, India \\
         \texttt{\href{mailto:adhyasuman30@gmail.com}{adhyasuman30@gmail.com}, \href{mailto:debarshi.sanyal@iacs.res.in}{debarshi.sanyal@iacs.res.in}}
}

\begin{document}
\maketitle
\begin{abstract}
Topic modeling is a widely used approach for analyzing and exploring large document collections. Recent research efforts have incorporated pre-trained contextualized language models, such as BERT embeddings, into topic modeling. However, they often neglect the intrinsic informational value conveyed by mutual dependencies between words. In this study, we introduce GINopic, a topic modeling framework based on graph isomorphism networks to capture the correlation between words. By conducting intrinsic (quantitative as well as qualitative) and extrinsic evaluations on diverse benchmark datasets, we demonstrate the effectiveness of GINopic compared to existing topic models and highlight its potential for advancing topic modeling.

\vspace{0.5em}
\hspace{.5em}\includegraphics[width=1.25em,height=1.25em]{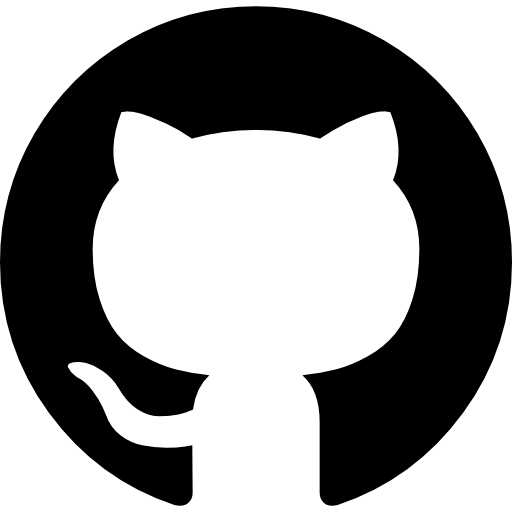}\hspace{.75em}\parbox{\dimexpr\linewidth-2\fboxsep-2\fboxrule}{\url{https://github.com/AdhyaSuman/GINopic}}
\vspace{-.5em}
\end{abstract}

\section{Introduction}
The rise in digital text data makes organizing them manually by theme increasingly difficult. Topic modeling plays a significant role here \cite{newman2010evaluating, boyd2017applications, adhya2022indian}, as it can uncover the underlying topics in documents in an unsupervised manner. In topic modeling, we assume that each document is a mixture of topics and these latent topics are also defined as distribution over the words.

\textbf{Motivation:} Recent approaches to neural topic modeling \citep{bianchi2020pre, bianchi2021cross, grootendorst2022bertopic} focus on the representation of the document as a sequence of words, which captures the contextual information. However, words in a document may be correlated to each other in a much more complex manner. So, why not explicitly consider these word dependency patterns while learning the topics? Several studies in the field of topic modeling delve into the representation of documents using graphs. In this context, nodes signify words, and edges depict relationships between words, such as syntax or semantic relations. For instance, in the case of short texts, the Graph Biterm Topic Model (GraphBTM) \citep{zhu2018graphbtm}, an extension of the Biterm Topic Model (BTM) \citep{yan2013btm}, represents word co-occurrence as a graph, with nodes representing words and weighted edges reflecting the counts of corresponding biterms. Despite GraphBTM's emphasis on capturing word dependencies, it has been reported to exhibit poor performance \citep{shen2021topic}. Additionally, its computational cost escalates with an expanding vocabulary, as it constructs a single graph using the entire vocabulary. In contrast, the Graph Neural Topic Model (GNTM) \citep{shen2021topic} employs a directed graph with word dependencies as edges between word nodes to incorporate semantic information from words in documents. However, GNTM considers word dependency solely by linking words within a small sliding window for a given document. This limitation makes it impossible to account for word dependencies that fall outside of that specific window. Furthermore, the computational complexity of generating document graphs increases with the length of the window.

\textbf{Approach:} To model the mutual dependency between words while addressing the existing issues of incorporation of document graphs into topic modeling, we developed a neural topic model that takes the word similarity graphs for each document, where the word similarity graph is constructed using word embeddings to capture the complex correlations between the words. These document graphs along with their unordered frequency-based text representation are then used as input. We have also used the Graph Isomorphism Network (GIN) to obtain the representation for each document graph. We have used GIN as it is provably the maximally powerful GNN under the neighborhood aggregation framework. It is as powerful as the Weisfeiler-Lehman graph isomorphism test \citep{xu2018how}.

\textbf{Contributions:} In summary, our work presents the following key contributions:

\begin{itemize}
    \item We introduce GINopic, a neural topic model that leverages a graph isomorphism network to enhance word correlations in topic modeling.    
    \item We perform a comprehensive analysis through quantitative, qualitative, and task-specific evaluations. Additionally, we visualize the latent spaces generated by our model to assess its capability to disentangle the latent representations of documents.
    \item We also conducted a sensitivity analysis for the selection of GIN among the GNNs and the choice of our graph construction methodology.
\end{itemize}

\section{Related Work}
Topic modeling processes extensive document collections efficiently, preserving key statistical relationships for tasks like classification, novelty detection, summarization, and similarity judgments. Traditional models like Latent Dirichlet Allocation (LDA) \citep{blei2003latent}, Probabilistic Latent Semantic Index (pLSI) \citep{hofmann2013probabilistic}, and Correlated Topic Model (CTM) \citep{lafferty2005correlated} use sampling-based algorithms or variational inference, but their design is limited by the need for careful selection, which limits the flexibility and scalability of model design.

Recent advancements in neural variational inference, particularly Auto Encoding Variational Bayes (AEVB) \citep{kingma2014auto}, simplify posterior computation. Neural Variational Document Model (NVDM) \citep{miao2016neural} is the pioneer VAE-based topic model. However, following the traditional topic models of applying Dirichlet-prior to the document-topic distribution becomes challenging due to the limitations of the reparametrization trick. Autoencoding Variational Inference For Topic Models (AVITM) \citep{srivastava2017autoencoding} resolves this by using Laplace approximation of the Dirichlet parameter with Gaussian parameters. CombinedTM \citep{bianchi2020pre} extends AVITM by incorporating sentence BERT embeddings alongside Bag-of-Words (BoW) representations. ZeroShotTM \citep{bianchi2021cross} further extends this approach, relying solely on SBERT embeddings, ignoring word co-occurrence relations in input documents.

Numerous contemporary methodologies incorporate Graph Neural Networks (GNNs) for topic modeling. In terms of the graph construction task, the Graph Biterm Topic Model (GraphBTM) \citep{zhu2018graphbtm} and the Graph Neural Topic Model (GNTM) \citep{shen2021topic} employ a moving window-based approach with a specified window length to model word co-occurrence relationships, necessitating careful window length selection. The \textit{graph topic model} \citep{zhou2020neural} constructs document graphs based on TF-IDF scores, capturing relationships with graph convolutions. \textit{Topic modeling with knowledge graph embedding} \citep{li2019integration} incorporates external knowledge graphs. The \textit{graph attention topic network} \citep{yang2020gat} addresses overfitting in probabilistic latent semantic indexing with amortized inference and word embeddings. The \textit{graph relational topic model} \citep{xie2021graph} explores document relationships using higher-order graph attention networks.

\section{Proposed Methodology}
Recognizing the challenges in topic modeling, we acknowledge the necessity of capturing semantic similarity among words in a document. Additionally, we note the importance of addressing the graph construction issue and obtaining unique representations for dissimilar document graphs. In response to these challenges, we have introduced the Graph Isomorphism Network-based neural topic model, abbreviated as GINopic. The following subsections provide a detailed explanation of the graph construction methodology, model framework, and objective function.

\subsection{Graph Construction}
Let $\mathcal{D}$ be defined as the set of all documents, $\mathcal{V}$ as the set of all words in the corpus such that $|\mathcal{V}|=V$, and $\mathcal{E} \in \mathbb{R}^{V \times \tau}$ as the word embeddings matrix such that its $i$-th row $\mathcal{E}_i \in \mathbb{R}^\tau$, corresponds to the word $w_i \in \mathcal{V}$. Now for a document $d \in \mathcal{D}$, which contains a subset of words from $\mathcal{V}$, specifically $V'$ words, we define its weighted undirected document graph $G_d$ as the adjacency matrix $A = (a_{ij})_{1 \leq i,j \leq V'}$, where the elements $a_{ij}$ are determined as follows:
\begin{align} \label{eq:adjacency}
    a_{ij} = \begin{cases}
              0 & \text{if } \operatorname{Sim}(\mathcal{E}_i, \mathcal{E}_j) < \delta  \\
              \operatorname{Sim}(\mathcal{E}_i, \mathcal{E}_j) & \text{otherwise}
            \end{cases}
\end{align}

\begin{figure}[!htbp]
    \centering
    \includegraphics[width=\linewidth]{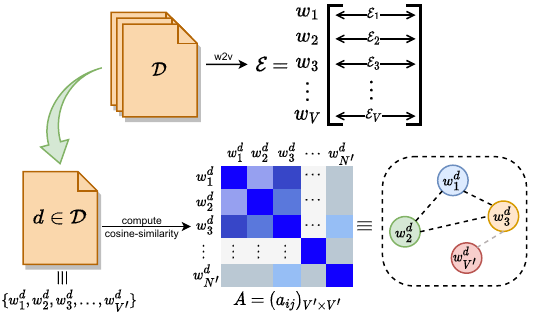}
    \caption{Graph construction methodology.}
    \label{fig:graph_construction}
\end{figure}

Here, $\operatorname{Sim}(\mathcal{E}_i, \mathcal{E}_i)$ represents the cosine similarity between the word embedding vectors $\mathcal{E}_i$ and $\mathcal{E}_j$. In Eq. \eqref{eq:adjacency}, $\delta$ is a threshold that indicates if the similarity score between two words is less than $\delta$ then there should not be any edge between them. The choice of this threshold is crucial, as opting for a lower value makes the connections in $G_d$ denser, consequently elevating computational complexity. Conversely, opting for a higher threshold value leads to a sparse document graph, a scenario also undesired. The optimal choice of $\delta$ depends on the type of corpus. To balance these factors, we consider $\delta$ as a hyperparameter to be tuned. 

\begin{figure*}
    \centering
    \includegraphics[width=\linewidth]{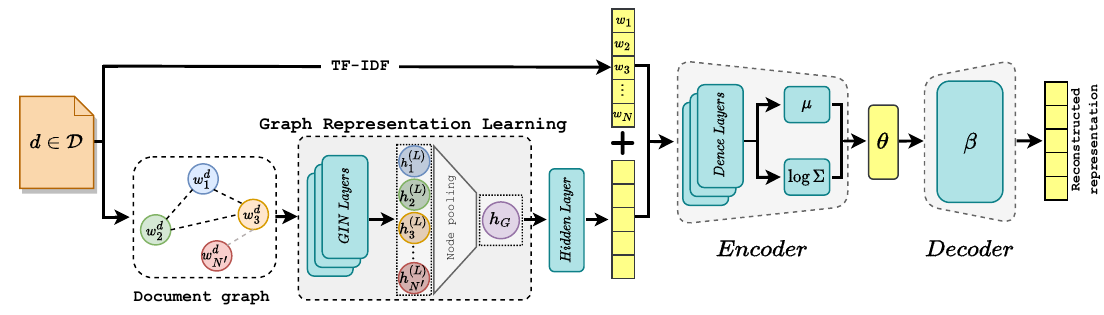}
    \caption{Proposed framework for GINopic model.}
    \label{fig:framework}
\end{figure*}

\subsection{Model Architecture}
The proposed model GINopic comprises a document graph representation learning network followed by an encoder which is followed by a decoder. The output of the graph representation learning network is concatenated with the TF-IDF representation of the input document before feeding into the encoder. The framework is shown in Fig. \ref{fig:framework} and a detailed description of these networks is described in the following.

\subsubsection{Graph Representation Learning} \label{sec:grl}
The Weisfeiler-Lehman (WL) test serves as a means to evaluate the isomorphism of two provided graphs. The graph representation learning module within the proposed model is designed to process document graphs as its input and produce a unique representation for each topologically distinct document graph, identified through the WL test. To model this injective mapping we have used the \textit{Graph Isomorphism Network} (GIN), known for its equivalent expressive power to the WL graph kernel \citep{nino2011wl}. GIN is theoretically proven as the most powerful GNN \citep{xu2018how}. Mathematically, the layer-wise propagation rule for GIN at layer $l+1$ is defined as follows:
\begin{align}
    h_i^{(l+1)} &= \MLP^{(l+1)} \bigg((1+\epsilon)h_i^{(l)} + \\ 
    & \qquad \operatorname{AGG}\Big( \big\{ \omega_{ji}h_j ^{(l)}, j \in N(i) \big\} \Big) \bigg) \nonumber
\end{align}
Here, $h_i^{(l)}$ represents the feature vector for the $i$-th node at layer $l$, $N(i)$ denotes the set of all neighbors for node $i$, $\omega_{ji}$ signifies the edge weight between the $i$-th and $j$-th nodes. The operator $\operatorname{AGG} (\cdot)$ stands for aggregation, and $\epsilon$ is a parameter that can be learned or a fixed scalar value close to zero. Furthermore, $\MLP^{(l+1)}$ represents the multi-layer perceptron for the $(l+1)$-th layer. After applying the $L$ number of GIN layers, the encoding of a node essentially captures its $L$-th order neighborhood's information. The detailed transformations are: $\big[ \gin(\tau, H) \rightarrow \bn \rightarrow \relu \rightarrow [\gin(H,H) \rightarrow \bn \rightarrow \relu ]^{L-2} \rightarrow \gin(H, \tau') \rightarrow \bn \big]$,  where $\gin(I,J)$ represents GIN layer with a MLP of input dimension $I$ and output dimension $J$, $H$ is the number of hidden units, $\bn$ is the batch normalization, and $\relu$ is the activation function. The final node embeddings of dimension $\tau'$ are then summed up to obtain the representation of the document graph as follows: $h_{G} = \sum_{i} h_i^{(L)}$.

\subsubsection{VAE framework} 
\textbf{Encoder Network:} The encoder network of GINopic, takes the combination of graph representation ($h_G \in \mathbb{R}^{\tau'} $) and TF-IDF representation ($x_{\operatorname{TFIDF}} \in \mathbb{R}^{V}$) of the input document. For this concatenation, $h_G$ is first scaled to the dimension same as of $x_{\operatorname{TFIDF}}$ and then concatenated with $x_{\operatorname{TFIDF}}$. Therefore, the resultant representation is $x = \operatorname{CONCAT}\big(f_W(h_G), x_{\operatorname{TFIDF}}\big)$, where $W \in \mathbb{R}^{V \times \tau'}$ is a matrix, representing linear transformation $f_W: \mathbb{R}^{\tau'} \rightarrow \mathbb{R}^{V}$ whose weights are to be learned.

Careful selection of the prior for our modeling assumption is crucial. In topic modeling, the Dirichlet distribution has been demonstrated \citep{Minno2009rethinking} as effective in assigning topic proportions for a given document. However, the reparametrization trick is limited to Gaussian distributions. To integrate the Dirichlet assumption into a VAE following the method proposed by \cite{srivastava2017autoencoding}, we used the Laplace approximation to the $\operatorname{Dir}(\alpha)$ distribution:
\begin{align*}
    \mu_{1\:k} &= \log{\alpha_k} - \frac{1}{K} \sum_i \log{\alpha_i} \\
    \Sigma_{1\:kk} &= \frac{1}{\alpha_k} \left(1-\frac{2}{K}\right) + \frac{1}{K^2} \sum_i ^K \frac{1}{\alpha_i}
\end{align*}
where, $\alpha_i$ is the $i$-th component of the $K$-dimensional Dirichlet's parameter, $\mu_{1\:k}$ is the $k$-th component of the vector $\mu_1 \in \mathbb{R}^K$ and $\Sigma_{1\:kk}$ is the $k$-th component of $\Sigma_1 \in \mathbb{R}^{K \times K}$, the diagonal covariance matrix. Given a prior distribution and the resultant input document representation vector $x$, the encoder outputs the posterior distribution $q_{\phi}(z \vert x) \equiv \mathcal{N} \left(\mu_0,  \Sigma_0 \right)$, where $\phi$ represents the weights of the encoder. The transformations in the encoder are: $\big[\linear(2V, H') \rightarrow \softplus \rightarrow [ \linear(H', H') \rightarrow \softplus ]^{L'-1} \rightarrow \drop(0.2) \big]$. This is followed by the two separate and similar transformations as follows: $\big[\linear(H', K) \rightarrow \bn \big]$ for $\mu_0$ and $\Sigma_0$ respectively. In these expressions, $V$ represents the vocabulary size, $H'$ and $L'$ represent the number of hidden units and hidden layers respectively, $\softplus$ is an activation function, and $\drop$ is a regularizer.

\textbf{Sampling Procedure:} A latent representation $z$ is stochastically sampled from the posterior distribution $q_{\phi}(z \vert x)$ using the reparameterization trick \citep{kingma2014auto} as $z = \mu_0 + \Sigma_0^{1/2} \odot \epsilon$. The symbol $\odot$ denotes the Frobenius inner product and $\epsilon \sim \mathcal{N}(0, 1)$. The obtained latent representation $z$ is then used as logit to a softmax function $\sigma(\cdot)$ in order to generate the document-topic distribution $\theta$ such that, $\theta = \Softmax(z)$.

\textbf{Decoder Network:} In the decoder, the topic-word matrix $\beta$ refers to the learnable weights of the decoder network. This matrix is utilized to reconstruct the word distribution $\hat{x}$ as: $\hat{x} = \Softmax \left( \beta^{\top} \theta \right)$ %= \Softmax \left( \beta^{\top} \Softmax(\mu_0 + \Sigma_0^{1/2} \odot \epsilon) \right)$.
Following \cite{srivastava2017autoencoding}, we relaxed the simplex constraint on $\beta$, which is empirically shown to produce better topic quality. The transformations of the decoder network are, $\big[ \linear(K, V) \rightarrow \bn  \rightarrow \operatorname{Softmax} \big]$, with $\Softmax$ employed in the output layer to generate the word distribution.

\subsection{Training Objective} 
The objective function for GINopic is the same as ELBO which needs to be maximized in order to maximize the log-likelihood of the input data distribution. The loss function we seek to minimize is defined as:
\begin{align}
   \label{eq:VAE_loss}
    \mathcal{L} &= \mathcal{L}_{\RL} + \mathcal{L}_{\KL} \\ \nonumber
    & \equiv - \mathbb{E}_{z \sim q_{\phi}(z \vert {x})} [p_{{\beta}}({x} \vert {z})]  + \operatorname{D}_{\KL}\left(q_{{\phi}}({z} \vert {x})\| p({z})\right) \nonumber
\end{align}

In the above expression, the first term ($\mathcal{L}_{\RL}$) represents the reconstruction loss, quantified by the cross-entropy between the predicted output distribution $\hat{x}$ and the input vector $x_{\operatorname{TFIDF}}$. On the other hand, the second term ($\mathcal{L}_{\KL}$) is the Kullback-Leibler (KL) divergence of the learned latent space distribution $q_{\phi}({z} \vert x)$ from the prior $p(z)$ of the latent space.

\section{Experimental Settings}\label{sec:expSetmpStudy}
We have conducted the experiments using OCTIS\footnote{\url{https://github.com/MIND-Lab/OCTIS}} \cite{terragni2021octis}, a comprehensive framework for comparing and optimizing topic models, available under the \href{https://github.com/MIND-Lab/OCTIS/blob/master/LICENSE}{MIT License.}

\subsection{Datsets}

\begin{table}[ht]
    \centering
    \begin{adjustbox}{width=.95\linewidth}
      \begin{tabular}{l c c c c c} \hline \hline
        %\toprule
        \multirow{2}{*}{\textbf{Dataset}} & \multirow{1}{*}{\textbf{\#Total}} & \multirow{1}{*}{\textbf{\#Tr}} & \multirow{1}{*}{\textbf{\#Ts/Va}} & \multirow{1}{*}{\textbf{Avg. Doc.}} & \multirow{2}{*}{\textbf{Labels}} \\
        & \multirow{1}{*}{\textbf{Docs}} & \multirow{1}{*}{\textbf{Docs}} & \multirow{1}{*}{\textbf{Docs}} & \multirow{1}{*}{\textbf{length}} & \\ \hline \hline
        
         \multirow{1}{*}{\textbf{20NG}} & $16309$ & 11415 & 2447 & \multirow{1}{*}{$48.020$} & $20$ \\ 
         \multirow{1}{*}{\textbf{BBC}} & $2225$ & 1557 & 334 & \multirow{1}{*}{$120.116$} & $5$ \\ 
         \textbf{SS} & $12270$ & $8588$ & $1841$ & $13.104$ & $8$ \\ 
         \textbf{Bio} & $18686$ & $13080$ & $2803$ & $7.022$ & $20$ \\
         \textbf{SO} & $15696$ & $10986$ & $2355$ & $5.106$ & $20$ \\ \hline \hline
         
      %\bottomrule
    \end{tabular}
    \end{adjustbox}
    \caption{Statistics of the used datasets. \label{tab:datasets}}
\end{table}

In the experiments, we utilized five publicly available datasets. Among these, \href{http://qwone.com/~jason/20Newsgroups/}{20NewsGroups (\textbf{20NG})} and BBC News (\textbf{BBC}) \citep{greene2006practical} datasets were already included in OCTIS in pre-processed formats. Additionally, we incorporated the SearchSnippets (\textbf{SS}), Biomedicine (\textbf{Bio}), and StackOverflow (\textbf{SO}) datasets \citep{qiang2022short} and pre-processed them. A detailed description of these datasets is mentioned in Appendix \ref{sec:ap_datasets} and the pre-processing steps are mentioned in Appendix \ref{sec:ap_preprocess}. Statistical descriptions of these datasets can be found in Table \ref{tab:datasets}. Each of these corpora was divided into training, validation, and testing sets, with a distribution ratio of 70\% for training, 15\% for validation, and 15\% for testing, where the training part is used to train the model, the validation part is only used for the \textbf{GNTM} to modify the learning rate accordingly and the test part is used to conduct the extrinsic evaluation of the models.

\subsection{Baselines}
We conducted a comparative analysis of the proposed model \textbf{GINopic} with the graph-based topic models, namely \textbf{GraphBTM} \citep{zhu2018graphbtm} and \textbf{GNTM} \citep{shen2021topic}. Unfortunately, for other graph-based topic models, we could not access their source code, making it impossible to include them in our comparison. Beyond the graph-based models, our evaluation extended to various well-known neural and traditional topic models, including \textbf{ECRTM} \cite{wu2023effective}, \textbf{CombinedTM} \citep{bianchi2020pre}, \textbf{ZeroShotTM} \citep{bianchi2021cross}, \textbf{ProdLDA} \citep{srivastava2017autoencoding}, \textbf{NeuralLDA} \citep{srivastava2017autoencoding}, \textbf{ETM} \citep{dieng2020topic}, \textbf{LDA} \citep{blei2003latent}, \textbf{LSI} \citep{dumais2004latent} and \textbf{NMF} \citep{zhao2017online}. A detailed description of the configurations of these baselines together with their implementation details can be found in Appendix \ref{sec:ap_config}.

\subsection{Hyperparameter Tuning}
\begin{table}[ht]
    \centering
    \begin{adjustbox}{width=\linewidth}
      \begin{tabular}{lccccc} \hline \hline
         \textbf{Hyperpramerts} & \textbf{20NG} & \textbf{BBC} & \textbf{SS} & \textbf{Bio} & \textbf{SO} \\ \hline \hline
         Graph construction threshold ($\delta$): & 0.4 & 0.3 & 0.2 & 0.05 & 0.1 \\
         Dim. of input node feature ($\tau$):  & 2048 & 256 & 1024 & 1024 & 64 \\
         \#GIN layers ($L$):  & 2 & 3 & 2 & 2 & 2 \\
         \#Hidden layers in MLP: & 1 & 1 & 1 & 1 & 1 \\
         Dim. of Hidden layers in MLP: & 200 & 50 & 50 & 200 & 300 \\
         Dim. of output node feature ($\tau'$): & 768 & 512 & 256 & 256 & 512\\
      \hline \hline
    \end{tabular}
    \end{adjustbox}
    \caption{Value of the hyperparameters of GINopic for each dataset. \label{tab:hyp_params}}
\end{table}
In GINopic, for a given dataset the hyperparameters that are tuned are mentioned in Table \ref{tab:hyp_params}. Here, the hyperparameter tuning was conducted on each dataset, maintaining a topic count equal to the number of labels for 50 epochs. To ensure a fair comparison we have also tuned the hyperparameters for GNTM. However, due to computational limitations, we are unable to fine-tune the hyperparameters for GraphBTM.

\begin{figure*}
    \centering
    \includegraphics[width=\linewidth]{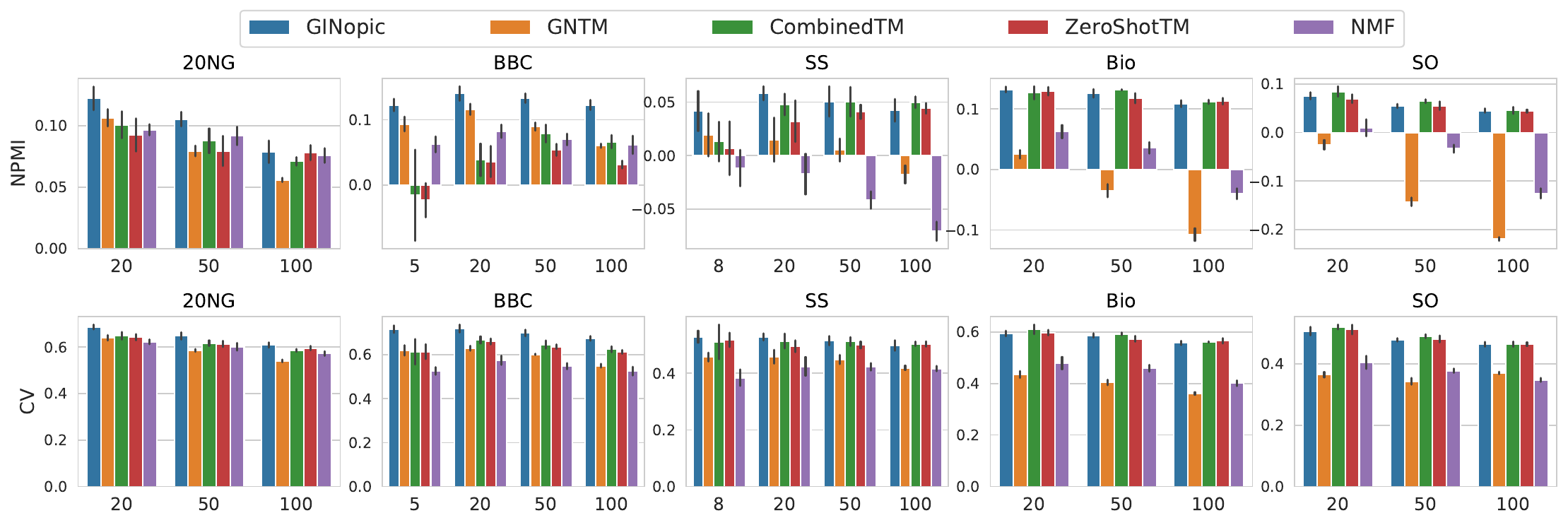}
    \caption{Topic coherence (NPMI and CV) scores for each topic count for top-5 topic models on five datasets.}
    \label{fig:quantitative}
\end{figure*}

\begin{table*}[ht]
    \centering
    \begin{adjustbox}{width=.9\linewidth}
      \begin{tabular}{l|cc|cc|cc|cc|cc} \hline \hline        
        \multirow{2}{*}{\textbf{Model}} & \multicolumn{2}{c|}{\textbf{20NG}} & \multicolumn{2}{c|}{\textbf{BBC}} & \multicolumn{2}{c|}{\textbf{SS}} & \multicolumn{2}{c|}{\textbf{Bio}} & \multicolumn{2}{c}{\textbf{SO}}\\ \cline{2-11}
                
        & \textbf{NPMI} & \textbf{CV} & \textbf{NPMI} & \textbf{CV} & \textbf{NPMI} & \textbf{CV} & \textbf{NPMI} & \textbf{CV} & \textbf{NPMI} & \textbf{CV}\\ \hline \hline

        \textbf{ECRTM} & -0.145 & 0.363 & -0.041 & 0.625 & -0.388 & 0.474 & -0.435 & 0.529 & -0.416 & 0.526 \\
        
        \textbf{CombinedTM} & 0.086 & 0.617 & 0.042 & 0.637 & 0.040 & 0.510 & \textbf{0.123} & 0.587 & \textbf{0.065} & 0.491 \\
        
        \textbf{ZeroShotTM} & 0.083 & 0.617 & 0.024 & 0.630 & 0.031 & 0.504 & 0.120 & 0.579 & 0.056 & 0.486 \\
        
        \textbf{ProdLDA} & 0.071 & 0.593 & 0.035 & 0.628 & -0.001 & 0.486 & 0.105 & 0.571 & 0.042 & 0.473 \\
        
        \textbf{NeuralLDA} & 0.045 & 0.500 & -0.065 & 0.472 & -0.114 & 0.400 & -0.061 & 0.435 & -0.177 & 0.407 \\
        
        \textbf{ETM} & 0.050 & 0.528 & 0.030 & 0.452 & -0.099 & 0.309 & -0.136 & 0.140 & -0.332 & 0.441 \\
        
        \textbf{LDA} & 0.069 & 0.562 & 0.049 & 0.518 & -0.165 & 0.376 & -0.118 & 0.392 & -0.174 & 0.345 \\
        
        \textbf{LSI} & -0.019 & 0.400 & -0.042 & 0.406 & -0.122 & 0.280 & -0.118 & 0.392 & -0.129 & 0.303 \\
        
        \textbf{NMF} & 0.088 & 0.599 & 0.069 & 0.543 & -0.035 & 0.412 & 0.019 & 0.446 & -0.050 & 0.377 \\
        
        \textbf{GraphBTM} & 0.017 & 0.605 & -0.173 & 0.484 & -0.322 & 0.444 & -0.398 & 0.519 & -0.451 & 0.558 \\
        
        \textbf{GNTM} & 0.081 & 0.588 & 0.090 & 0.600 & 0.005 & 0.445 & -0.039 & 0.400 & -0.129 & 0.359 \\ \hline
        
        \textbf{GINopic} & \textbf{0.102} & \textbf{0.647} & \textbf{0.130} & \textbf{0.701} & \textbf{0.048} & \textbf{0.517} & \textbf{0.123} & \textbf{0.589} & 0.059 & \textbf{0.493} \\ \hline \hline     
    \end{tabular}
    \end{adjustbox}
    \caption{Comparison of topic models on five datasets. For each metric and each topic model, we mention the mean scores over topic counts $\{20, 50, 100\} \cup \{k_{gold}\}$.\label{tab:NPMI_CV_results}}
\end{table*}

\begin{table*}[ht]
    \centering
    \begin{adjustbox}{width=\linewidth}
      \begin{tabular}{l|ccc|ccc|ccc|ccc|ccc} \hline \hline        
        \multirow{2}{*}{\textbf{Model}} & \multicolumn{3}{c|}{\textbf{20NG}} & \multicolumn{3}{c|}{\textbf{BBC}} & \multicolumn{3}{c|}{\textbf{SS}} & \multicolumn{3}{c|}{\textbf{Bio}} & \multicolumn{3}{c}{\textbf{SO}}\\ \cline{2-16}
                
        & \textbf{IRBO}  & \textbf{wI-M} & \textbf{wI-C} & \textbf{IRBO}  & \textbf{wI-M} & \textbf{wI-C} & \textbf{IRBO}  & \textbf{wI-M} & \textbf{wI-C} & \textbf{IRBO}  & \textbf{wI-M} & \textbf{wI-C} & \textbf{IRBO}  & \textbf{wI-M} & \textbf{wI-C}\\ \hline \hline

        \textbf{ECRTM} & \textbf{0.998} & \textbf{0.473} & 0.852 & \textbf{0.999} & 0.454 & 0.848 & \textbf{1.000} & 0.442 & 0.839 & \textbf{1.000}	& 0.433	& 0.838	& \textbf{1.000} & 0.382 & 0.825 \\
        
        \textbf{CombinedTM} & 0.988 & 0.468 & \textbf{0.895} & 0.978 & 0.442 & 0.888 & 0.993 & 0.45 & 0.888 & 0.983 & 0.443 & 0.887 & 0.985 & 0.392 & 0.878\\
        
        \textbf{ZeroShotTM} & 0.986	& 0.467	& 0.894  & 0.964 & 0.435 & 0.887 & 0.99 & 0.448 & 0.888 & 0.983 & 0.445 & 0.885 & 0.985 & 0.393 & 0.879\\
        
        \textbf{ProdLDA} & 0.990 & 0.469 & \textbf{0.895} & 0.975 & 0.44 & 0.888 & 0.994 & 0.45 & 0.888 & 0.987 & 0.446 & 0.888 & 0.977 & 0.394 & 0.878 \\
        
        \textbf{NeuralLDA} & 0.989 & 0.466 & 0.892 & 0.984	& 0.444	& 0.887 & 0.997 & 0.453 & 0.887 & 0.996 & 0.452 & 0.888 & 0.979 & 0.390 & 0.875\\
        
        \textbf{ETM} & 0.802 & 0.37 & 0.87 & 0.802 & 0.354 & 0.874 & 0.647 & 0.294 & 0.867 & 0.344 & 0.138 & 0.843 & 0.490 & 0.187 & 0.842 \\
        
        \textbf{LDA} & 0.981 & 0.462 & 0.893 & 0.947 & 0.424 & 0.885 & 0.988 & 0.447 & 0.886 & 0.991 & 0.446 & 0.886 & 0.913 & 0.390 & 0.875 \\
        
        \textbf{LSI} & 0.925 & 0.429 & 0.887 & 0.869 & 0.385 & 0.879 & 0.845 & 0.382 & 0.881 & 0.991 & 0.399 & 0.881 & 0.927 & 0.337 & 0.868\\
        
        \textbf{NMF} & 0.975 & 0.458 & 0.892 & 0.966 & 0.432 & 0.886 & 0.978 & 0.443 & 0.887 & 0.988 & 0.443 & 0.887 & 0.984 & 0.388 & 0.876 \\
        
        \textbf{GraphBTM} & 0.971 & 0.462 & 0.852 & 0.986 & 0.448 & 0.846 & 0.947 & 0.421 & 0.836 & 0.924 & 0.427 & 0.837 & 0.958 & 0.374 & 0.821\\
        
        \textbf{GNTM} & 0.984 & 0.461 & 0.852 & 0.983 & 0.444 & 0.845 & 0.995 & \textbf{0.454} & 0.846 & 0.999 & 0.455 & 0.845 & 0.949 & 0.406 & 0.831 \\ \hline
        
        \textbf{GINopic} & 0.989 & 0.468 & \textbf{0.895} & 0.992 & \textbf{0.457} & \textbf{0.893} & 0.998 & \textbf{0.454} & \textbf{0.889} & 0.983 & \textbf{0.462} & \textbf{0.888} & 0.986 & \textbf{0.497} & \textbf{0.879} \\ \hline \hline      
      %\bottomrule
    \end{tabular}
    \end{adjustbox}
    \caption{Comparison of topic models on five datasets. For each metric and each topic model, we mention the mean scores over topic counts $\{20, 50, 100\} \cup \{k_{gold}\}$.\label{tab:Diversity_results}}
\end{table*}

\section{Results and Discussions}
We categorize our findings into the following sections: (1) quantitative evaluation (Section \ref{sec:quantative}), (2) extrinsic evaluation (Section \ref{sec:extrinsic}), (3) qualitative evaluation (Section \ref{sec:qualitative}), (4) latent space visualization (Section \ref{sec:latentSpaceVis}), and (5) sensitivity analysis (Section \ref{sec:sensitivity}).

\subsection{Quantitative Evaluation} \label{sec:quantative}
In the quantitative evaluation, we have evaluated the topic models based on the generated topic quality measured by coherence and diversity metrics. To measure the topic coherence we have used \textbf{Normalized Pointwise Mutual Information} (\textbf{NPMI}) \citep{lau2014machine} and \textbf{Coherence Value} (\textbf{CV}) \citep{roder2015exploring}. NPMI is commonly utilized \cite{adhya2022improving} as a surrogate for human judgment of topic coherence, although some researchers also employ CV, despite its known \href{https://palmetto.demos.dice-research.org/}{issues}. We measure the diversity of topics using \textbf{Inverted Rank-Biased Overlap} (\textbf{IRBO}) \cite{bianchi2020pre}, \textbf{Word Embedding-based Inverted Rank-Biased Overlap - Match} (\textbf{wI-M}), and \textbf{Word Embedding-based Inverted Rank-Biased Overlap - Centroid} (\textbf{wI-C}) \cite{terragni2021word}. Higher values of NPMI, CV, IRBO, wI-C, and wI-M indicate better performance. These metrics are elaborately discussed in Appendix \ref{sec:ap_TopicEval}.

\textbf{Experimental Setup:} For a given dataset we run all the models by varying the topic count in $\{20, 50, 100\} \cup \{k_{gold}\}$ where $k_{gold}$ stands for the \textit{golden topic count} which is the number of ground-truth labels in the dataset (since they are available for the datasets we used). The values of $k_{gold}$ for \textbf{20NG}, \textbf{BBC}, and \textbf{SS} are 20, 5, and 8, respectively. For the robustness of the results, we have reported the mean value over 5 random runs for a given model, a given dataset, and a given topic count.

\textbf{Findings:} We present coherence scores for all models across datasets in Table \ref{tab:NPMI_CV_results}. Notably, GINopic achieves the highest coherence scores (both NPMI and CV) across most datasets, except for the \textbf{SO} dataset where it ranks second in NPMI score, following CombinedTM. However, GINopic still leads in CV score for the \textbf{SO} dataset. To provide a comprehensive comparison, we focus on the top 5 models based on their NPMI scores across all datasets. Figure \ref{fig:quantitative} shows the mean and standard deviation of NPMI and CV scores for each topic count. The results establish the consistent superior performance of GINopic compared to existing models. In terms of diversity, Table \ref{tab:NPMI_CV_results} displays all three diversity scores. GINopic achieves the highest wI-M and wI-C diversity scores across most datasets, except for the \textbf{20NG} dataset where its wI-M score is comparable to ECRTM's highest score. ECRTM exhibits the highest IRBO scores across all datasets due to its embedding clustering regularization approach, despite its poor coherence scores indicating ineffective topic representation learning. IRBO scores of GINopic are also competitive, being close to the highest score across all datasets.

\subsection{Extrinsic Evaluation} \label{sec:extrinsic}
We have also incorporated an extrinsic task to assess the performance of the topic models, specifically by evaluating their predictive capabilities in a document classification task. 

\textbf{Experimental Setup:} Our datasets include category labels for each document. We trained all models on the training subset of a particular dataset to generate $k_{gold}$ topics. The resulting $k_{gold}$-dimensional \textit{document-topic} vector serves as a representation of the document. A linear support vector machine is then trained on these representations, and model performance on the test subset is reported. We calculate the average accuracy over five runs for each dataset and present the scores in Table \ref{tab:Doc_classification}.

\textbf{Findings:} Table \ref{tab:Doc_classification} shows that GINopic attains the highest accuracy across all datasets, except for the 20NG dataset where it secures the second-highest accuracy, with the GNTM closely edging ahead. 

\begin{table}[ht]
    \centering
    \begin{adjustbox}{width=\linewidth}
      \begin{tabular}{ l  c  c  c  c  c} \hline \hline
        \textbf{Model} & \makecell{\textbf{20NG}} & \makecell{\textbf{BBC} } & \makecell{\textbf{SS}} & \makecell{\textbf{Bio}} & \makecell{\textbf{SO}}\\
        \hline \hline
        \textbf{ECRTM} & 0.411 & 0.816 & 0.492 & 0.361 & 0.457\\
        \textbf{CombinedTM} & 0.397 & 0.796 & 0.706 & 0.493 & 0.715\\
        \textbf{ZeroShotTM} & 0.385 & 0.817 & 0.698 & 0.501 & 0.687\\
        \textbf{ProdLDA} & 0.385 & 0.752 & 0.662 & 0.489 & 0.674\\
        \textbf{NeuralLDA} & 0.297 & 0.575 & 0.464 & 0.376 & 0.403\\
        \textbf{ETM} & 0.370 & 0.754 & 0.496 & 0.083 & 0.072\\
        \textbf{LDA} & 0.428 & 0.798 & 0.440 & 0.364 & 0.412\\
        \textbf{LSI} & 0.329 & 0.337 & 0.343 & 0.402 & 0.660\\
        \textbf{NMF} & 0.350 & 0.785 & 0.415 & 0.437 & 0.708\\
        \textbf{GraphBTM} & 0.052 & 0.231 & 0.224 & 0.060 & 0.050 \\
        \textbf{GNTM} & \textbf{0.449} & 0.806 & 0.222 & 0.049 & 0.053 \\  \hline
        \textbf{GINopic} & 0.441 & \textbf{0.888} & \textbf{0.713} & \textbf{0.566} & \textbf{0.785} \\ \hline \hline     
    \end{tabular}
    \end{adjustbox}
    \caption{Average accuracy scores in the document classification task for all the models trained with topic count $k_{gold}$ for all five datasets. \label{tab:Doc_classification}}
\end{table}

\subsection{Latent Space Visualization} \label{sec:latentSpaceVis}
We have further examined the latent space generated by GINopic. In topic modeling, documents are projected into a lower-dimensional latent (topic) space.

\begin{figure}[!htbp]
    \centering
    \includegraphics[width=\linewidth]{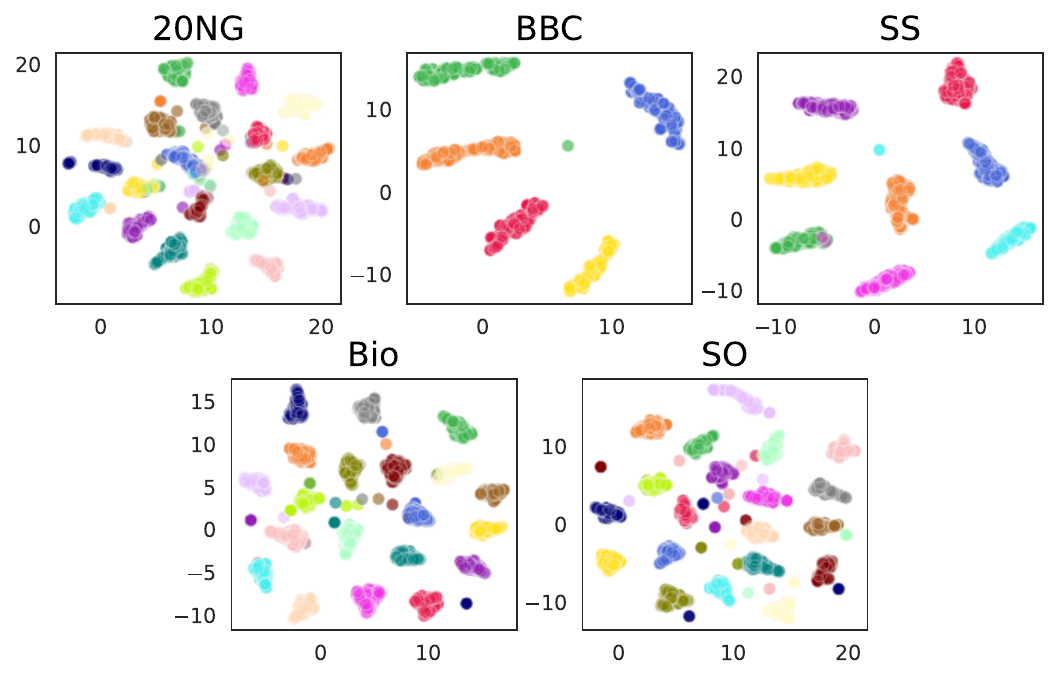}
    \caption{Latent space visualization for GINopic model across all five datasets.}
    \label{fig:umap}
\end{figure}

\textbf{Experimental Setup:} To visualize the latent space, we have trained GINopic for the topic count of $k_{gold}$ associated with each of the five datasets. Following the training phase, we captured the document-topic distribution for each document. We applied the \textit{Uniform Manifold Approximation and Projection} (UMAP) technique, a robust dimensionality reduction method \citep{mcinnes2018umap-software}. UMAP transformed the $k_{gold}$-dimensional document-topic distribution into a two-dimensional representation, making it possible to visualize. Each document was assigned to a cluster based on its topic distribution vector $\vec{\theta}$, where the cluster was determined by selecting the topic with the highest probability. Figure \ref{fig:umap} illustrates the clusters obtained for each dataset.

\textbf{Findings:} The disentanglement of clusters is depicted in Figure \ref{fig:umap} for each dataset. Notably, the clarity of disentanglement is more pronounced in the \textbf{BBC} and \textbf{SS} datasets compared to the other three datasets \textbf{20NG}, \textbf{Bio}, and \textbf{SO}. This difference can be attributed to the greater challenge of disentangling 20 different labels in the \textbf{20NG}, \textbf{Bio}, and \textbf{SO} datasets, as opposed to the \textbf{BBC} and \textbf{SS} datasets with fewer distinct labels.

\subsection{Qualitative Evaluation} \label{sec:qualitative}

\begin{table*}[ht]
\centering
\begin{adjustbox}{width=\linewidth}
  \begin{tabular}{l l} \hline \hline
  \textbf{Model} & \multicolumn{1}{c}{\textbf{Topics}} \\ \hline \hline
  
  \multirow{3}{*}{\textbf{ECRTM}} 
        & \textbf{turkish}, \textbf{soviet}, \textbf{bullet}, \textbf{minority}, \textbf{population}, \textbf{burn}, \textbf{jewish}, cold, prepare, joke \\
        
        & draft, \textbf{baseball}, \textbf{game}, \textbf{shot}, blue, luck, stupid, programming, basically, \textbf{score} \\ 
        
        & \textbf{bike}, \textbf{car}, \textbf{controller}, button, camera, strategy, win, black, atheism, attribute\\ \hline
        
    \multirow{3}{*}{\textbf{CombinedTM}} 
        & \textbf{german}, publish, \textbf{genocide}, \textbf{turkish}, \textbf{muslim}, \textbf{armenian}, book, representative, \textbf{european}, century \\
        
        & \textbf{team}, \textbf{hockey}, \textbf{season}, \textbf{game}, draft, expansion, \textbf{ticket}, \textbf{play}, year, ice \\ 
        
        & \textbf{car}, \textbf{engine}, \textbf{tire}, \textbf{bike}, \textbf{ride}, \textbf{brake}, good, problem, buy, \textbf{mile} \\ \hline
         
    \multirow{3}{*}{\textbf{ZeroShotTM}}
        & \textbf{greek}, \textbf{turkish}, \textbf{minority}, \textbf{genocide}, state, \textbf{muslim}, \textbf{soviet}, \textbf{armenian}, \textbf{israeli}, \textbf{struggle} \\
        
        & ranger, \textbf{hockey}, \textbf{playoff}, \textbf{team}, \textbf{game}, devil, king, pen, wing, period \\ 
        
        & \textbf{motorcycle}, \textbf{bike}, clean, wave, \textbf{ride}, \textbf{wheel}, tip, \textbf{mirror}, remove, replace \\ \hline
         
    \multirow{3}{*}{\textbf{ProdLDA}}
         & \textbf{arab}, \textbf{israeli}, \textbf{religious}, \textbf{people}, \textbf{religion}, \textbf{jewish}, solution, \textbf{territory}, understanding, \textbf{land} \\
         
         & year, fund, money, spend, program, \textbf{player}, private, \textbf{team}, job, good \\ 
         
         & eat, food, \textbf{car}, problem, \textbf{engine}, \textbf{brake}, stone, weight, pain, day \\ \hline
         
    \multirow{3}{*}{\textbf{NeuralLDA}}
        & \textbf{army}, \textbf{muslim}, \textbf{genocide}, \textbf{international}, \textbf{turkish}, \textbf{village}, \textbf{armenian}, \textbf{population}, organize, enter \\
        
        & \textbf{goal}, \textbf{win}, \textbf{score}, \textbf{play}, \textbf{wing}, \textbf{penalty}, \textbf{playoff}, \textbf{team}, \textbf{pass}, \textbf{game} \\ 
        
        & \textbf{front}, clean, \textbf{ride}, foot, \textbf{bike}, bar, \textbf{engine}, pull, weight, remove \\ \hline
        
    \multirow{3}{*}{\textbf{ETM}}
        & \textbf{armenian}, \textbf{people}, \textbf{turkish}, \textbf{village}, \textbf{kill}, \textbf{genocide}, \textbf{woman}, live, \textbf{soldier}, \textbf{jewish} \\
        
        & good, year, \textbf{win}, \textbf{game}, back, \textbf{play}, make, post, line, \textbf{goal} \\ 
        
        & \textbf{bike}, \textbf{engine}, mission, orbit, temperature, \textbf{car}, earth, space, planet, solar \\ \hline
         
    \multirow{3}{*}{\textbf{LDA}}
        & \textbf{war}, \textbf{jewish}, \textbf{israeli}, \textbf{land}, \textbf{country}, \textbf{arab}, \textbf{peace}, \textbf{territory}, \textbf{force}, \textbf{attack} \\
        
        & double, trade, \textbf{game}, \textbf{hockey}, final, \textbf{team}, star, \textbf{playoff}, king, regular \\ 
        
        & \textbf{bike}, \textbf{ride}, hate, advice, bank, \textbf{motorcycle}, weight, good, instruction, surrender \\ \hline

    \multirow{3}{*}{\textbf{LSI}}
        &  \textbf{turkish}, drive, \textbf{war}, \textbf{armenian}, \textbf{russian}, \textbf{government}, \textbf{secret}, \textbf{military}, \textbf{power}, \textbf{jewish} \\
        
        & year, car, scsi, love, bit, client, \textbf{team}, server, call, \textbf{player} \\ 
        
        & access, \textbf{engine}, \textbf{power}, kill, database, word, \textbf{bus}, attack, disk, card \\ \hline

    \multirow{3}{*}{\textbf{NMF}}
        & \textbf{kill}, \textbf{woman}, time, \textbf{soldier}, start, \textbf{child}, back, leave, \textbf{armenian}, \textbf{man}  \\
        
        & power, \textbf{play}, government, constitution, \textbf{team}, control, level, individual, idea, zone \\ 
        
        & \textbf{car}, \textbf{engine}, price, buy, \textbf{bike}, \textbf{mile}, \textbf{ride}, make, \textbf{driver}, \textbf{tire} \\ \hline

    \multirow{3}{*}{\textbf{GraphBTM}}
        &  \textbf{armenian}, \textbf{afraid}, neighbor, clock, \textbf{soldier}, \textbf{turkish}, floor, \textbf{soviet}, \textbf{beat}, arrive \\
        
        & \textbf{game}, \textbf{score}, car, engine, \textbf{play}, \textbf{goal}, \textbf{season}, \textbf{playoff}, \textbf{shot}, \textbf{player} \\
        
        & \textbf{tire}, \textbf{bike}, connector, ide, \textbf{brake}, scsi, cable, \textbf{car}, \textbf{rear}, \textbf{engine} \\ \hline

    \multirow{3}{*}{\textbf{GNTM}}
        & \textbf{israeli}, \textbf{arab}, \textbf{jewish}, policy, \textbf{land}, \textbf{territory}, \textbf{area}, peace, \textbf{human}, \textbf{population} \\
        
        & \textbf{team}, \textbf{game}, \textbf{play}, \textbf{player}, \textbf{win}, year, good, call, point, time \\
        
        & \textbf{tire}, \textbf{oil}, \textbf{brake}, \textbf{bike}, \textbf{paint}, weight, corner, air, \textbf{lock}, \textbf{motorcycle} \\ \hline

    \multirow{3}{*}{\textbf{GINopic}}
        & \textbf{genocide}, \textbf{muslim}, \textbf{armenian}, \textbf{massacre}, \textbf{turkish}, \textbf{population}, \textbf{kill}, \textbf{government}, \textbf{troop}, \textbf{war} \\
        
        & \textbf{team}, \textbf{win}, \textbf{score}, \textbf{baseball}, \textbf{game}, \textbf{player}, \textbf{hockey}, \textbf{playoff}, \textbf{goal}, \textbf{play} \\ 
        
        & \textbf{car}, \textbf{bike}, \textbf{ride}, \textbf{brake}, \textbf{light}, \textbf{tire}, \textbf{engine}, \textbf{lock}, side, \textbf{mile} \\ \hline \hline

  %\bottomrule
\end{tabular}
\end{adjustbox}
\caption{Some representative topics extracted from the \textbf{20NG} dataset with a topic count of 100. Relevant terms within each topic are emphasized in \textbf{bold}. \label{tab:topics_20NG}}
\end{table*}

Since topic models operate as unsupervised methods, it is recommended to assess their performance not solely relying on automated estimates of topic coherence but also through manual evaluation of the topics, as emphasized by \cite{hoyle2021automated, adhya2023improving}.

\textbf{Experimental Setup:} We conducted a qualitative analysis of the topics, utilizing the \textbf{20NG} dataset and training all models with the golden topic count i.e. $k_{gold}=20$. The results appear in Table \ref{tab:topics_20NG}. Note that the table exhibits aligned topics, wherein the first topic listed for one model is similar to the first topic for every other model, and the same goes for the rest of the topics, following the alignment method proposed by \cite{adhya2023neural}. Additionally, words closely associated with a given topic are highlighted in \textbf{bold}.

\textbf{Findings:} In Table \ref{tab:topics_20NG}, we showcase three topics: ``Armenian genocide", ``Sports", and ``Automobile" related. Across these distinct topics, GINopic consistently generates more correlated words compared to other models. This observation is supported by the consistently higher number of \textbf{bold} words for each topic in GINopic, indicating stronger word correlations than the other models.

\subsection{Sensitivity Analysis} \label{sec:sensitivity}

\subsubsection{Choice of the Graph Neural Network}
\begin{figure*}
    \centering
    \includegraphics[width=.9\linewidth]{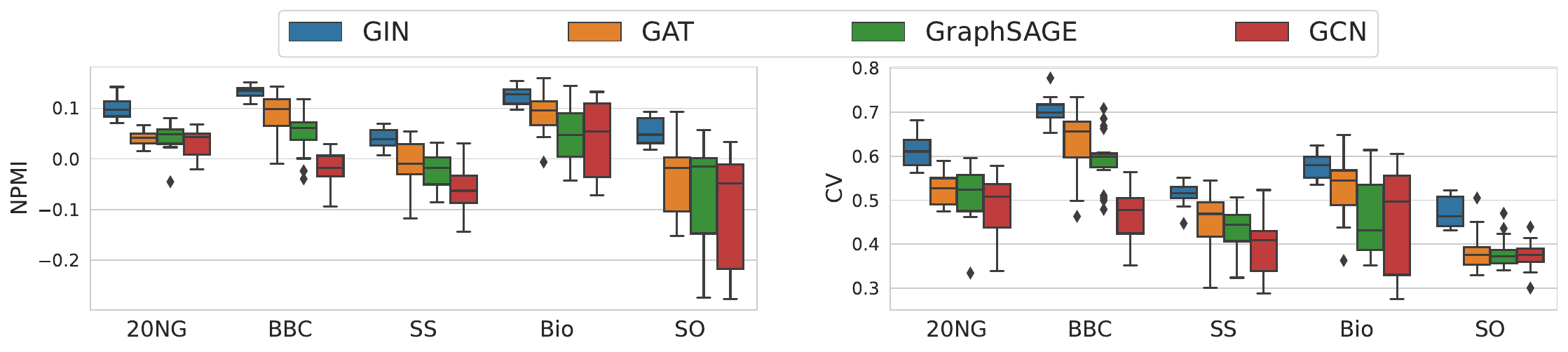}
    \caption{Box plot of topic coherence (NPMI and CV) scores incorporating GIN, GAT, GraphSAGE, and GCN in GINopic on five datasets.}
    \label{fig:Sensitivity_GIN}
\end{figure*}
To empirically check the effectiveness of GIN over other GNNs in our model, we substitute GIN with Graph Attention Network (GAT) \citep{petar2018graph}, Graph SAmple and aggreGatE (GraphSAGE) \citep{hamilton2017inductive} and Graph Convolutional Network (GCN) \citep{kipf2017semi}.

\textbf{Experimental Setup:} We trained our proposed model on the five datasets, adjusting the topic count within the set $\{20, 50, 100\} \cup {k_{gold}}$. To ensure a fair comparison, we maintained consistent parameter values across all models, aligning them with those of GINopic.

\textbf{Findings:} In Figure \ref{fig:Sensitivity_GIN}, a box plot is presented, illustrating the NPMI and CV scores derived from five random runs for each model across the five datasets. The results indicate that the GIN-incorporated model consistently outperforms other GNN-based models across all datasets in terms of both the coherence measures.

\subsubsection{Choice of the Graph Construction Threshold ($\delta$)}
\begin{figure}
    \centering
    \includegraphics[width=\linewidth]{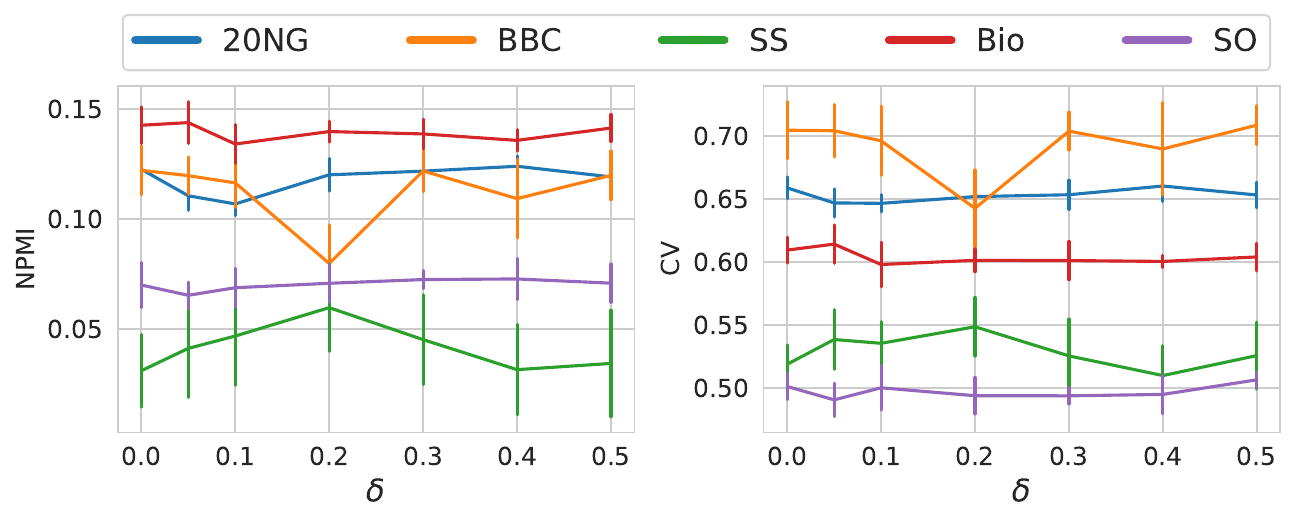}
    \caption{Coherence (NPMI and CV) scores for each dataset by varying the threshold ($\delta$) value in $\{0.0, 0.05, 0.1, 0.2, 0.3, 0.4, 0.5\}$.}
    \label{fig:Sensitivity_delta}
\end{figure}
We have examined how the graph construction threshold $\delta$, as specified in Eq. \eqref{eq:adjacency}, influences model performance and training time.

\textbf{Experimental Setup:} Given a dataset, we have trained our model for the corresponding $k_{gold}$ number of topics by varying the value of $\delta$ over $\{0.0, 0.05, 0.1, 0.2, 0.3, 0.4, 0.5\}$. We reported the mean and standard deviation of the coherence scores (NPMI and CV) over 5 random runs in Figure \ref{fig:Sensitivity_delta}.

\textbf{Findings:} Figure \ref{fig:Sensitivity_delta} illustrates the optimal threshold values that maximize coherence scores (NPMI and CV) for each dataset. This threshold signifies that if the similarity between two nodes in a document graph falls below it, no edge connects those nodes. Moreover, increasing the $\delta$ value results in a sparser document graph, leading to reduced training time. Table \ref{tab:delta_results} provides details of the dataset-wise optimal threshold ($\delta$) values and the corresponding percentage reductions in training time from that with the $\delta$ value of 0.0. Thus, by tuning $\delta$, we improve the coherence scores and simultaneously reduce the training time.

\begin{table}[ht]
    \centering
    \begin{adjustbox}{width=\linewidth}
      \begin{tabular}{ l  c  c  c  c  c} \hline \hline
        \textbf{Value} & \makecell{\textbf{20NG}} & \makecell{\textbf{BBC} } & \makecell{\textbf{SS}} & \makecell{\textbf{Bio}} & \makecell{\textbf{SO}}\\
        \hline \hline
        \textbf{Optimal threshold} ($\mathbf{\delta}$) & 0.4 & 0.3 & 0.2 & 0.05 & 0.1\\
        \textbf{Reduction} ($\mathbf{\%}$) & $154.27\%$ & $266.72\%$ & $16.71\%$ & $0.29\%$ & $1.06\%$\\ \hline \hline     
    \end{tabular}
    \end{adjustbox}
    \caption{Optimal threshold ($\delta$) value along with the percentage of training time reduction for all five datasets. \label{tab:delta_results}}
\end{table}

\section{Conclusion}
We have introduced GINopic, a neural topic model based on a graph isomorphism network, and evaluated its performance on five widely used datasets for assessing topic models. Across the majority of our experiments, GINopic consistently exhibits superior topic coherence and diversity compared to other competitive topic models from the literature. Manual evaluation of selected topics further confirms that GINopic generates more coherent topics than alternative models. In extrinsic evaluations, GINopic generally outperforms existing models across all the datasets, except for the \textbf{20NG} dataset. We utilized visualizations of the latent space generated by GINopic to assess its clustering disentanglement capability. Sensitivity analysis demonstrates the impact of graph construction threshold values on the performance and training time of GINopic. Additionally, we highlight the effectiveness of GIN over other graph neural networks in our topic model.

\section*{Limitations}
This paper focuses solely on utilizing word similarity for constructing document graphs. However, there exist alternative methods for constructing document graphs, such as incorporating dependency parse graphs. Future extensions of this work could explore capturing diverse word dependencies and integrating them to construct a multifaceted document graph.

\section*{Ethics Statement}
The topic words presented in Table \ref{tab:topics_20NG} depict the output of the topic models trained on the \textbf{20NG} dataset. The authors have no intention to cause harm or offense to any community, religion, country, or individual.

\appendix

\section{Data Overview}
\subsection{Dataset Descriptions} \label{sec:ap_datasets}
Datasets used in experiments:
\begin{enumerate}
    \item \textbf{20NewsGroups (20NG)} dataset comprising $16,309$ pre-processed documents from $20$ different newsgroups posts. Each document is labeled with its corresponding category type.
    
    \item \textbf{BBC News (BBC)} \cite{greene2006practical} dataset consists of $2,225$ news articles from BBC. Documents are categorized into $5$ different classes: \textit{tech}, \textit{business}, \textit{entertainment}, \textit{sports}, and \textit{politics}.
    
    \item \textbf{SearchSnippets (SS)} \cite{qiang2022short} is derived from predefined phrases across $8$ domains, this dataset is constructed from web search transactions. The domains include {business}, {computers}, {culture-arts}, {education-science}, {engineering}, {health}, {politics-society}, and {sports}.

    \item \textbf{Biomedicine (Bio)} \cite{qiang2022short} makes use of the challenge data delivered on BioASQ's official website.

    \item \textbf{StackOverflow (SO)} \cite{qiang2022short} The dataset is released on Kaggle.com. The raw dataset contains 3,370,528 samples from July 31st, 2012 to August 14, 2012. Here, the dataset randomly selects 20,000 question titles from 20 different tags.
\end{enumerate}

The initial two datasets, 20NG, and BBC, are available on OCTIS\footnote{\url{https://github.com/MIND-Lab/OCTIS}}. As for the remaining three datasets SS, Bio, and SO, we have pre-processed them using the method detailed in Section \ref{sec:ap_preprocess}.

\begin{table}[!htbp]
\centering
\begin{adjustbox}{width=\linewidth}
  \begin{tabular}{c l c c} \hline \hline
    \textbf{\#No.} & \textbf{Label} & \textbf{\#Docs} & \textbf{\%Docs} \\ \hline \hline
    1. & misc.forsale & 861 & 5.28 \\
    2. & comp.windows.x & 883 & 5.41 \\
    3. & soc.religion.christian & 920 & 5.64 \\
    4. & talk.religion.misc & 521 & 3.19 \\
    5. & rec.autos & 822 & 5.04 \\
    6. & sci.med & 866 & 5.31 \\
    7. & talk.politics.misc & 689 & 4.22 \\
    8. & talk.politics.mideast & 828 & 5.08 \\
    9. & sci.electronics & 867 & 5.32 \\
    10. & rec.sport.hockey & 843 & 5.17 \\
    11. & rec.sport.baseball & 787 & 4.83 \\
    12. & talk.politics.guns & 808 & 4.95 \\
    13. & sci.crypt & 883 & 5.41 \\
    14. & comp.sys.mac.hardware & 838 & 5.14 \\
    15. & comp.sys.ibm.pc.hardware & 891 & 5.46 \\
    16. & comp.graphics & 836 & 5.13 \\
    17. & comp.os.ms-windows.misc & 828 & 5.08 \\
    18. & alt.atheism & 689 & 4.22 \\
    19. & sci.space & 856 & 5.25 \\
    20. & rec.motorcycles & 793 & 4.86 \\ \hline \hline
\end{tabular}
\end{adjustbox}
\caption{\textbf{20NG} labels with corresponding document counts and percentage of documents. \label{tab:20NG_labels}}
\end{table}

\begin{table}[!htbp]
\centering
\begin{adjustbox}{width=.9\linewidth}
  \begin{tabular}{c l c c } \hline \hline
    \textbf{\#No.} & \textbf{Label} & \textbf{\#Docs} & \textbf{\%Docs} \\ \hline \hline
    1. & tech & 401 & 18.02 \\
    2. & business & 510 & 22.92 \\
    3. & entertainment & 386 & 17.35 \\
    4. & sport & 511 & 22.97 \\
    5. & politics & 417 & 18.74 \\ \hline \hline
\end{tabular}
\end{adjustbox}
\caption{\textbf{BBC} labels with corresponding document counts and percentage of documents. \label{tab:BBC_labels}}
\end{table}

\begin{table}[!htbp]
\centering
\begin{adjustbox}{width=.9\linewidth}
  \begin{tabular}{c l c c } \hline \hline
    \textbf{\#No.} & \textbf{Label} & \textbf{\#Docs} & \textbf{\%Docs} \\ \hline \hline
    1. & business & 2652 & 21.61 \\
    2. & computers & 2177 & 17.74 \\
    3. & culture-arts & 1499 & 12.22 \\
    4. & education-science & 1498 & 3.01 \\
    5. & engineering & 1491 & 12.15 \\
    6. & health & 1411 & 12.21 \\
    7. & politics-society & 1173 & 9.56 \\
    8. & sports & 369 & 11.5 \\ \hline \hline
\end{tabular}
\end{adjustbox}
\caption{\textbf{SS} labels with corresponding document counts and percentage of documents. \label{tab:SS_labels}}
\end{table}

\subsection{Preprocessing} \label{sec:ap_preprocess}
Using OCTIS, we convert each document to lowercase, remove the punctuations, lemmatize it, filter the vocabulary with the most frequent 2000 terms, filter words with less than 3 characters, and filter documents with less than 3 words.

\section{Baseline Configurations} \label{sec:ap_config}
We reproduced all baseline models by following the guidance provided in their original papers and utilizing codes from either the original sources or from OCTIS. Specifically, for \textbf{CombinedTM} \citep{bianchi2020pre}, \textbf{ZeroShotTM} \citep{bianchi2021cross}, \textbf{ProdLDA} \citep{srivastava2017autoencoding}, \textbf{NeuralLDA} \citep{srivastava2017autoencoding}, \textbf{ETM} \citep{dieng2020topic}, \textbf{LDA} \citep{blei2003latent}, \textbf{LSI} \citep{dumais2004latent}, \textbf{NMF} \citep{zhao2017online}, we employed the implementation from OCTIS with default parameter values. For \textbf{GraphBTM}\footnote{\url{https://github.com/valdersoul/GraphBTM}} \citep{zhu2018graphbtm}, \textbf{GNTM}\footnote{\url{https://github.com/SmilesDZgk/GNTM}} \citep{shen2021topic}, and ECRTM\footnote{\url{https://github.com/BobXWu/ECRTM}} \cite{wu2023effective} we utilized the official source codes. Hyperparameter optimization was performed for GNTM on each dataset, and the values are detailed in Table \ref{tab:gntm_hyp_params}. However, hyperparameter optimization for GBTM is computationally intensive, likely due to its exhaustive consideration of all words in the vocabulary when constructing the graph.
\begin{table}[ht]
    \centering
    \begin{adjustbox}{width=\linewidth}
      \begin{tabular}{lccc} \hline \hline
         \textbf{Hyperpramerts} & \textbf{20NG} & \textbf{BBC} & \textbf{SS} \\ \hline \hline
         Temperature for STGS: & 0.6 & 0.6 & 0.7\\
         Window size for graph construction:  & 3 & 2 & 10 \\
      \hline \hline
    \end{tabular}
    \end{adjustbox}
    \caption{Hyperparameter values for GNTM on each dataset. \label{tab:gntm_hyp_params}}
\end{table}

\section{Coherence Metrics} \label{sec:ap_TopicEval}
Coherence matrices are used to compute the relevance of the top words within topics. The NPMI topic coherence for a given topic $\beta_k$ with $n$ top words is calculated as follows:

\begin{equation*}
    \operatorname{NPMI}(\beta_k) =  \frac{1}{\binom{n}{2}}\sum_{i=1}^n \sum_{j=i+1}^n \frac{\log{\frac{p(w_i, w_j) + \epsilon}{p(w_i)p(w_j)}}}{- \log{\left(p(w_i, w_j)+\epsilon\right)}}
\end{equation*}

Here, $p(w_i, w_j)$ is the probability of co-occurrence of words $w_i$ and $w_j$ in a boolean sliding window in topic $k$, and $p(w_i)$ and $p(w_j)$ represent the probability of the individual words' occurrence in topic $k$. $\epsilon$ is a small positive constant to prevent zero in the $\log(\cdot)$ function. NPMI ranges from $-1$ (words never co-occur) to $+1$ (they always co-occur). CV is computed using an indirect cosine measure along with NPMI scores over a boolean sliding window. In our experiments, we consider the top 10 words for each topic (i.e., $n=10$) to compute NPMI and CV scores.

\section{Diversity Metrics}
Topic diversity quantifies the uniqueness of generated topics. To measure the topic diversity we have used three following metrics: 
\begin{enumerate*}[label=(\roman*)] \item IRBO \cite{bianchi2021cross}, \item wI-M \cite{terragni2021word}, \item wI-C \cite{terragni2021word}. \end{enumerate*}
The IRBO gives $0$ for identical topics and $1$ for completely dissimilar topics. Suppose we are given a collection $\aleph$ of $T$ topics where each topic is a list of words such that the words at the beginning of the list have a higher probability of occurrence (i.e., are more important or more highly ranked) in the topic. Then, the IRBO score of the topics is defined as,
\begin{equation*}
    \IRBO(\aleph) = 1 - \frac{\sum_{i=2}^T \sum_{j=1}^{i-1} \RBO(l_i, l_j)}{n} 
\end{equation*}
where $n = {\binom{T}{2}}$ is the number of pairs of lists, and $\RBO(l_i, l_j)$ denotes the standard Rank-Biased Overlap between two ranked lists $l_i$ and $l_j$ \cite{webber2010similarity}. IRBO allows the comparison of lists that may not contain the same items, and in particular, may not cover all items in the domain. Two lists (topics) with overlapping words receive a smaller IRBO score when the overlap occurs at the highest ranks of the lists than when it occurs at lower ranks. IRBO is implemented in OCTIS.

\section{Computing Infrastructure} \label{sec:ap_Computing_Infrastructure}
Our experiments were run on a workstation with Intel\textsuperscript{\textregistered} Xeon\textsuperscript{\textregistered} W-1350 @ 3.30GHz, 6 Cores, 12 Threads, 16.0 GB RAM, NVIDIA RTX A4000 GPU, CUDA Version: 12.2 and Ubuntu 22.04 operating system.

\end{document}